\documentclass[letterpaper, 10 pt, conference]{ieeeconf}  

\IEEEoverridecommandlockouts                              

\usepackage{amsmath} 
\usepackage{amssymb}  
\usepackage[backend=bibtex,bibstyle=ieee,citestyle=numeric-comp,maxbibnames=3,maxcitenames=3,doi=false,url=false,eprint=false]{biblatex}
\usepackage{algorithm}
\usepackage{algorithmic}
\usepackage{tabularx}
\usepackage{multirow}
\usepackage{graphicx} 
\usepackage{amsfonts} 
\usepackage[dvipsnames]{xcolor}

\usepackage{subcaption}
\usepackage{url}
\usepackage{hyperref}

\title{\LARGE \bf
KI-PMF: Knowledge Integrated Plausible Motion Forecasting
}

\author{Abhishek Vivekanandan$^{1}$, Ahmed Abouelazm$^{1}$, Philip Schörner$^{1}$, J. Marius Zöllner$^{1,2}$
\thanks{$^{1}$ FZI Research Center for Information Technology, 76131 Karlsruhe, Germany.
        {\tt\small \{vivekana, abouelazm, schoerner, zoellner\} @fzi.de }}%
\thanks{$^{2}$ Karlsruhe Institute of Technology (KIT), Germany.}%
}

\begin{document}
\maketitle
\thispagestyle{empty}
\pagestyle{empty}

\begin{abstract}
The accurate prediction of surrounding traffic actors' movements is vital for the large-scale safe deployment of autonomous vehicles. Existing motion forecasting methods primarily aim to minimize prediction error by optimizing a loss function, which can sometimes lead to physically infeasible predictions or states that violate external constraints. This paper proposes a method that integrates explicit knowledge priors, allowing a network to forecast future trajectories that comply with both the vehicle's kinematic constraints and the driving environment's geometry. This is achieved by introducing a non-parametric pruning layer, and learnable attention layers to incorporate the defined knowledge priors. The proposed method aims to satisfy reachability guarantees and thus prevent off-road predictions common to a neural network in the motion forecasting realm. By conditioning the network to adhere to physical laws, we can achieve accurate and safe predictions, which are crucial for maintaining the safety and efficiency of autonomous vehicles in real-world settings.

\end{abstract}

\section{Introduction}
In complex driving environments, the ability to accurately and reliably predict the short-term trajectories of nearby traffic participants is crucial for upholding both safety and efficiency \cite{li2015real} in driving operations. These traffic participants often display multi-modal intentions, which translate into a range of plausible actions. When attempting to model this prediction space using Deep Neural Networks, the inherent multi-modality from observations introduces uncertainty, potentially leading to predictions of implausible future states. In scenarios where safety is paramount, such inaccuracies can result in critical failures, as the network may not effectively incorporate prior knowledge, a concern highlighted in \cite{bahari2022vehicle, cruisecrash}. The challenges outlined here are significant barriers to deploying these predictive models in autonomous driving systems, where ensuring safety is of utmost importance.

\begin{figure}[t]
  \begin{subfigure}{0.49\columnwidth}
  \includegraphics[width=\textwidth]{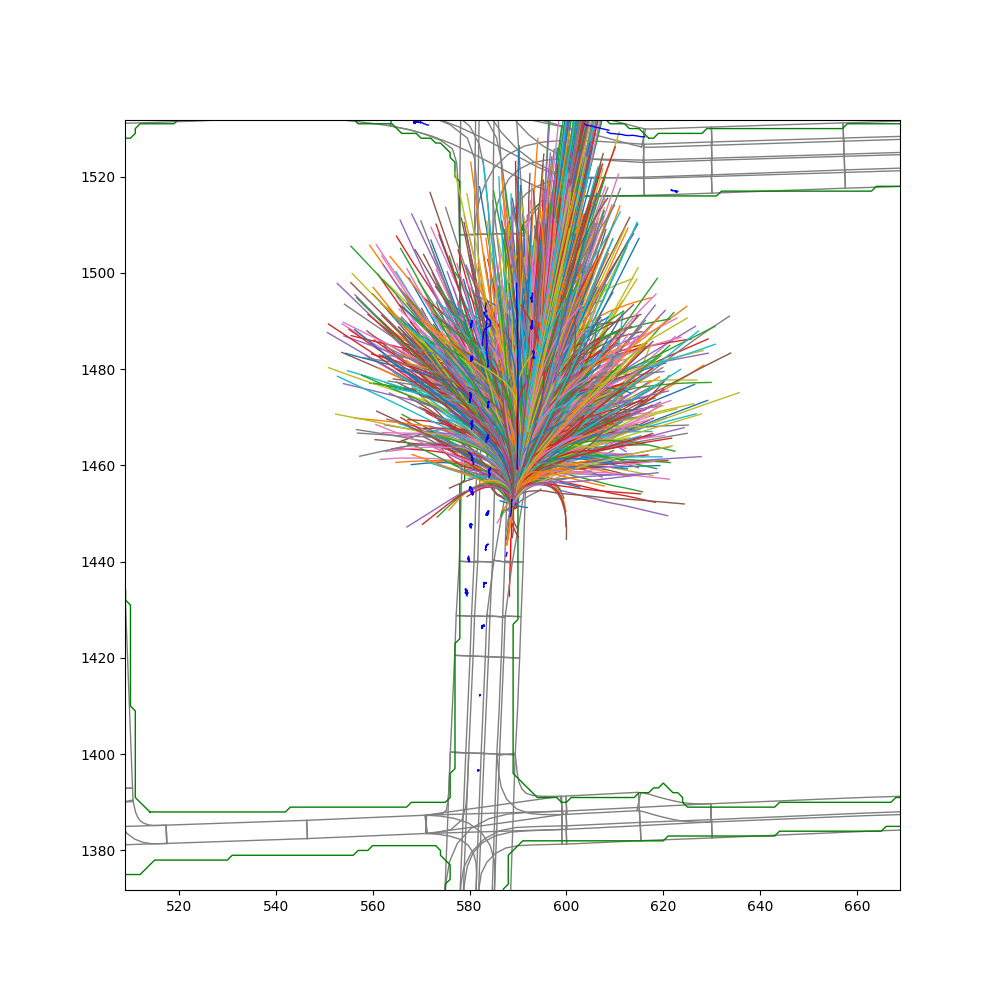}
  \caption{Fig.1a}
  \end{subfigure}
  \hfill
  \begin{subfigure}{0.49\columnwidth}
  \includegraphics[width=\textwidth]{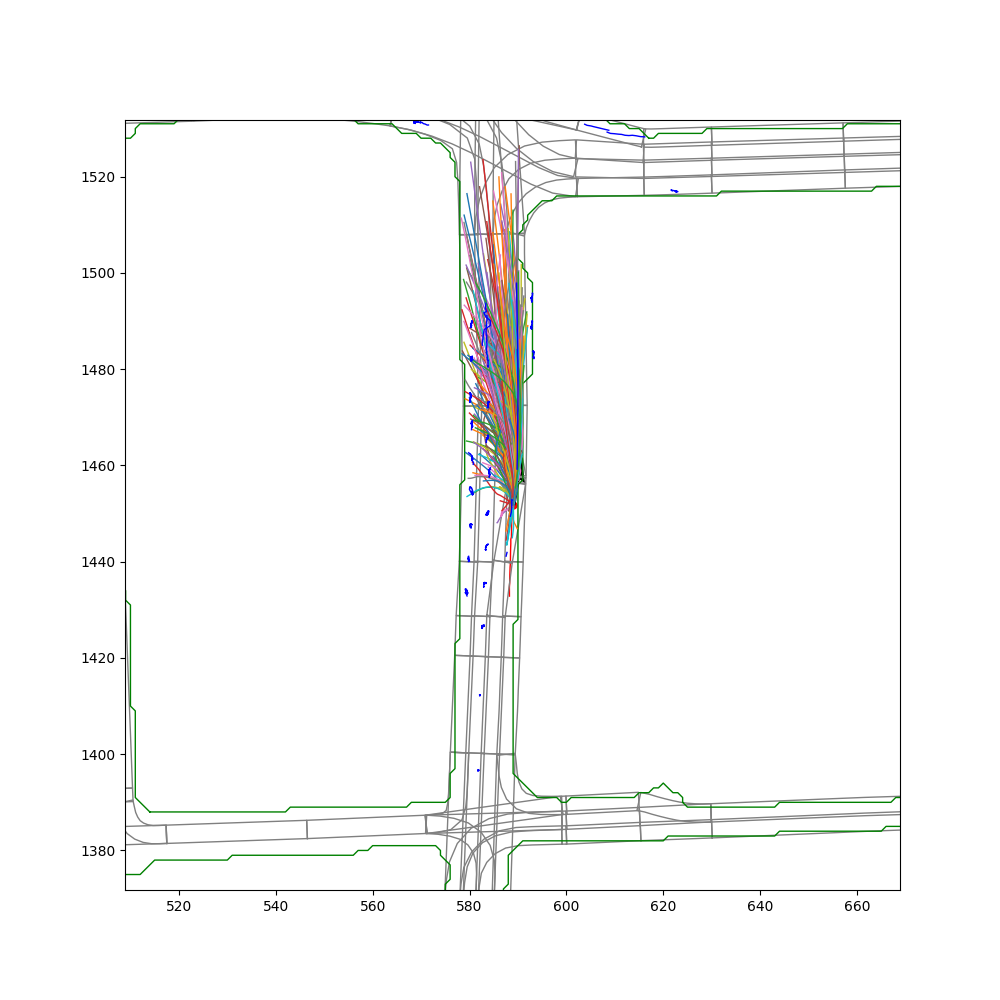}
  \caption{Fig.1b} 
  \end{subfigure} 
  \begin{subfigure}{0.49\columnwidth} 
  \includegraphics[width=\textwidth]{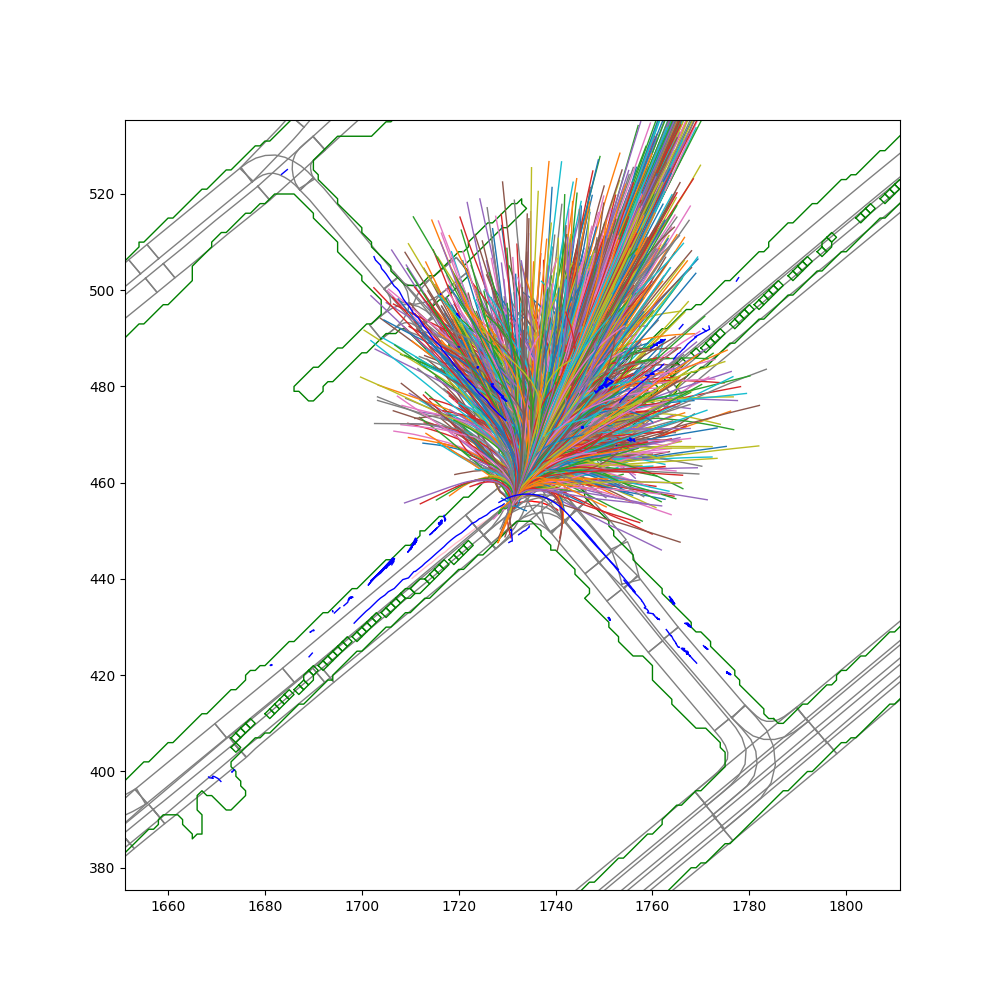} 
  \caption{Fig.1c} 
  \end{subfigure}  
  \hfill 
  \begin{subfigure}{0.49\columnwidth} 
  \includegraphics[width=\textwidth]{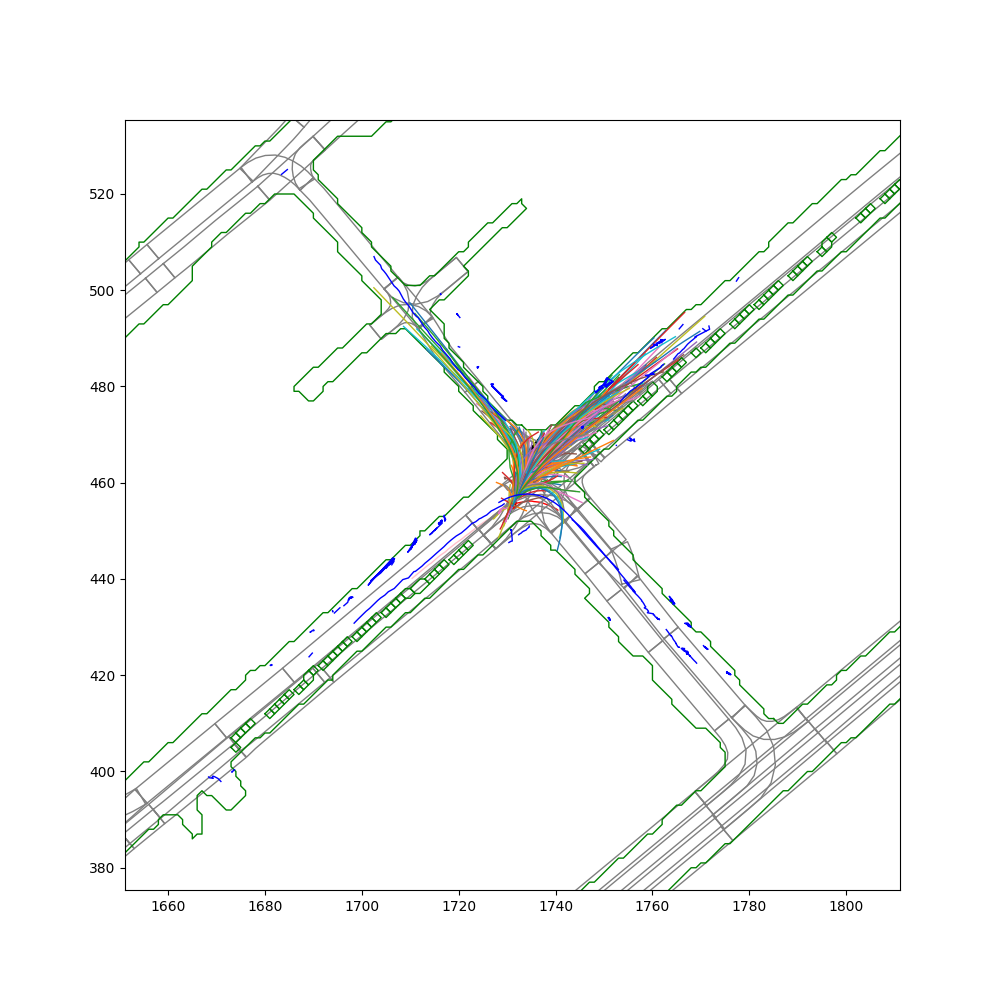} 
  \caption{Fig.1d} 
  \end{subfigure}
  \caption{The left column represents the two scenes with unrefined trajectories injected into the scene. The right column represents the impact of our refinement layer, where we refine the original trajectory set with priors, allowing us to avoid predicting in the off-road areas.}
  \end{figure}
In this work, we introduce a unique multi-stage framework that incorporates explicit knowledge, referred to as constraints or priors, into a learning-based trajectory forecasting model. This ensures plausible motion forecasting for traffic actors. Our model is composed of two stages. The first stage is a deterministic, non-parametric refinement layer that employs explicit environmental and kinematic constraints to compute plausible trajectories. We frame the multi-modal trajectory prediction problem as a classification problem, constructing future reachable states as a set of trajectories, as inspired by the works of Phan-Minh et al. \cite{covernet}. This output representation method enables us to incorporate non-holonomic vehicle constraints \cite{walsh1994stabilization} into the learning paradigm, which would otherwise be challenging. Environmental constraints, derived from HD maps M, are encoded as polygons for a Point-in-Polygon search, further refining the trajectories and eliminating off-road predictions. The advantage of this non-parametric search is that it ensures the proposals are interpretable. In the second phase of our approach, the model utilizes the refined trajectory proposals and integrates them with designated goal positions \cite{gu2021densetnt} to construct a viable representation space, delineating potential regions for actor movement. This fusion of lanes and trajectories enables the network to learn reachability-based features from a scene at a global level for the target actor.

The key contributions of this work include:
\begin{itemize}
\item A two-stage trajectory generation and prediction approach with a constraint-based refinement layer, ensuring feasibility and safety by eliminating off-road trajectories and implausible states.
\item The use of multi-headed attention layers to learn from the interactions between feasible trajectories and corresponding goal-based reachable lanes.
\item The computation of intensity calculations to demonstrate the real-time capabilities of the refinement layer.
\end{itemize}

\section{Related Work}

\textbf{Scene representation and encoding} are important parts of a motion forecasting stack, enabling the learning of a latent representation of the scene from various input modalities. Several methodologies, such as those presented in \cite{covernet,bansal2018chauffeurnet,chai2019multipath}, utilize a rasterized image of the scene as input, which is then processed through CNNs to extract latent features. The integration of stacked rasterized images into the multimodal prediction architecture is demonstrated in the works of \cite{cui2019multimodal}, where different heads are employed for regressing trajectories as waypoints and classification heads for estimating the softmax probabilities of the trajectories. A sensor-independent feature representation in a spatially discretized grid is proposed in \cite{schorner2019grid}, where a 2D CNN encoder-decoder architecture is used to predict the environment's future states. 

Recent methodologies, as presented in \cite{vectornet,lanegcn,zhao2021tnt,schmidt2022crat,grimm2023holistic}, have shifted their focus towards utilizing vector-based representations of the scene. This shift is prompted by the impractical computational costs and constraints associated with the receptive field of CNNs utilized for processing rasterized images. 

\textbf{Existing Limitations} include the limited generalization to unseen situations and lack of safety guarantees in trajectory forecasting networks were investigated in \cite{bahari2022vehicle}. In this study, adversarial scenes are created by making slight modifications to scenes sourced from the Argoverse dataset \cite{chang_argoverse_2019}. Established methodologies, including WIMP \cite{khandelwal2020if} and LaneGCN \cite{lanegcn}, initially trained on the Argoverse dataset with low off-road rate in the original scenes, undergo re-evaluation on these adversarially generated scenes. The outcomes reveal a noteworthy occurrence of off-road or physically implausible trajectories, manifesting in approximately $60\%$ of the scenes. The empirical observations derived from this study, highlighting the limited generalization of deep learning approaches, promote a towards the explicit integration of prior knowledge. The integration of priors aims to enhance the reliability of predicted trajectories by seeking safety guarantees in the forecasting process. Although prior knowledge integration and plausibility-based checks are done for applications related to 3D object detectors \cite{vivekanandan_plausibility_2023}, our work is the first to extend it to motion forecasting problems.

In the realm of motion forecasting, the integration of \textbf{vehicular motion constraints} has been explored through various studies, such as those by Schörner et al. \cite{schorner2020optimization}, Zhu et al. \cite{zhu2020safe}, and others, which combine machine learning techniques with optimization to ensure the creation of safe trajectories. However, these approaches often apply feasibility checks as a subsequent step rather than during the learning phase, leading to potential inefficiencies in computational resource use. Historically, motion forecasting has been dominated by regression techniques that incorporate constraints through vehicle dynamics modeling, as seen in works by Cui et al. \cite{cui_dkm} and others \cite{pfvtp_harshayu}. These methods effectively prevent infeasible trajectories based on vehicle turn-rate but lack mechanisms to prevent off-road predictions due to missing environmental constraint

Although Covernet \cite{covernet} compiles future trajectories of actors from various scenes into a set of trajectories, it also allows for the creation of an alternative set by applying valid longitudinal and lateral accelerations to a vehicle's kinematic model throughout the prediction horizon. A subset of these trajectories, known as a covering set \cite{branicky2008path}, is selected by optimizing to maximize coverage over the trajectory sets. A network is then trained to classify over the covering set for a given scene representation in a supervised fashion. However, this method has several shortcomings, such as the lack of physical feasibility guarantees for the compiled trajectories, and its dependence on rasterized inputs, which hinders its use in real-time applications. Moreover, the methodology does not address the issue of off-road predictions. The classification task is also made more complex by the fact that the covering set is not integrated into the network's input, leading to a form of blind learning. RESET \cite{schmidt2023reset} sought to overcome some of Covernet's limitations by using LaneGCN as a backbone and adopting a vector-based scene representation. It also moved away from the optimization problem-solving approach for generating covering sets, favoring a metric-driven method. Nevertheless, RESET still grapples with the fundamental issues of blind classification and the challenge of off-road predictions.

While PRIME \cite{song2022learning} successfully integrates learning and trajectory generation within a single pipeline, it creates trajectories in Frenet space by altering the terminal lateral distance and velocity of a path from the vehicle's initial pose to its reachable goal positions. This approach necessitates an additional pre-processing step for each reachable path, which involves projecting all surrounding actors onto the Frenet space generated by that path, thereby adding computational overhead. Moreover, the proposed network does not incorporate HD map information. Finally, the trajectories generated in Frenet space are based on a smooth assumption and demonstrate limited effectiveness in capturing complex maneuvers.

In this work, we propose a novel approach that prevents off-road predictions and guarantees physically valid trajectories through the usage of a non-parametric refinement layer. The guarantees are derived by including physical and outer constraints into the trajectory generation and consequently propagated into the learning components of the network for reliable predictions.
\section{Methodology}

\subsection{Problem formulation}
The objective of motion forecasting in autonomous driving deals with predicting the future possible trajectories of actors in a given scenario. The past trajectories of surrounding actors are assumed to be tracked through an existing perception system. Given a target actor $a_{tar}$ for prediction, trajectory forecasting utilizes not only the past observed trajectories of $a_{tar}$ but also of other actors in its vicinity $a_{other}$ over a finite time window $t_p$ for accurate prediction of its future trajectory over a prediction horizon $t_h$. A past trajectory $T_{p,i}$ of an actor $a_i$ is represented by a sequence of states, as shown in equation \ref{eq:previous_traj}, where $s_i^t$ represents the centerpoint location $(x, y)$ of the actor $i$ at a time $t$ in a Cartesian coordinate frame. The set of all actors' past trajectories is noted as $\Gamma_{p}$.

\begin{equation}
    T_{p,i} = \left \{ s_i^{-t_p}, s_i^{-t_p + 1}, s_i^{-t_p + 2}, ..., s_i^{0} \right \}
    \label{eq:previous_traj}
\end{equation}

Similarly, the future predicted trajectory $T_{h,i}$ of a target actor $a_i$ is a sequence of predicted center point coordinates, as seen in equation \ref{eq:future}. Such that the set of all actors' future trajectories is $\Gamma_{h}$.
\begin{equation}
    T_{h,i} = \left \{ s_{i}^{1}, s_{i}^{2}, s_{i}^{3}, ..., s_{i}^{t_h}\right \}
    \label{eq:future}
\end{equation}

\begin{figure*}[htb]
    \vspace{0.2cm}
    \centering
    \includegraphics[width=0.80\linewidth, keepaspectratio]{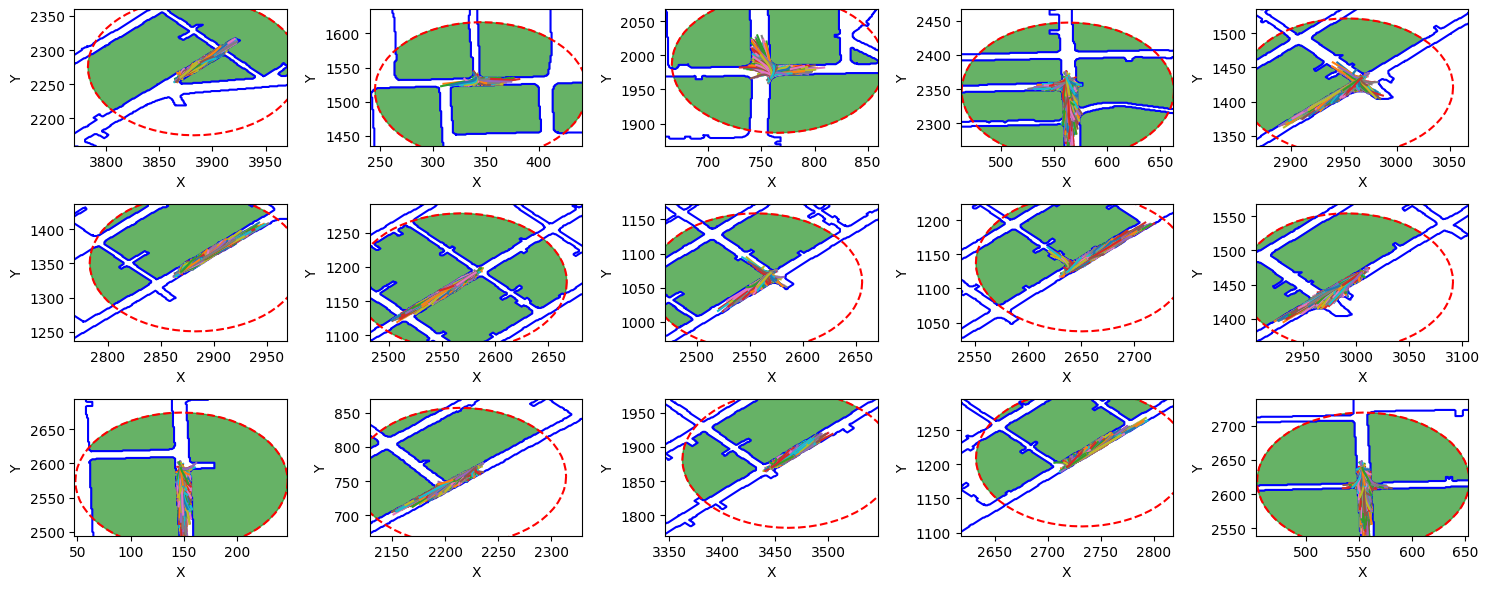}
    \caption{The \textcolor{PineGreen}{green} area delineates the local buffer polygon, instrumental in diminishing the spatial extent of the principal polygonal; non-drivable regions. The \textcolor{blue}{blue} linear entities correspond to the polygonal boundaries. For illustrative purposes, a circular buffer, depicted by a \textcolor{red}{red} dotted circumference, is employed. Subsequently, the peripheries of the green polygons undergo collision detection analysis to expedite the trajectory computation within the polygonal confines The x and y-axis are represented in the city coordinates system.}
    \label{fig:local_buffer_construction}
\end{figure*}

Moreover, we utilize HD map information, as illustrated in Figure~\ref{fig:local_buffer_construction}, where each lane within a query window is discretized and encoded as a set of centerline $(x, y)$ points in a coordinate system relative to the prediction target agent $a_{tar}$. The components involved in the architecture will be discussed in more detail in the following sections. By incorporating HD map information, the model can better understand the surrounding environment and enhance the accuracy of trajectory predictions

\subsection{First Stage: Integration of Prior Knowledge}

An initial trajectory set is generated with a bagging algorithm, similar to the one proposed in Covernet \cite{covernet}, on Argoverse dataset \cite{chang_argoverse_2019} with pre-defined coverage $\epsilon \simeq 2m$. The coverage $\epsilon$ represents the maximum distance between two neighboring trajectories in the set. While the bagging algorithm is proficient in identifying a diverse set of trajectories, we augment it by imposing vehicular kinematic constraints. This augmentation ensures that the resultant trajectory set $\Gamma_{set}$ is smooth and kinematically feasible.

The first stage takes this trajectory set as input along with lanes' centerlines, converted to a vector representation from the HD map, to produce feasible trajectories that respect the road topology. 

\textbf{Coordinate transformation} The motion of target actors and HD map are represented in a city coordinate system \cite{chang_argoverse_2019} whereas the trajectory sets are in Cartesian coordinates. To define a common relative representation for the networks to learn from, we use the curvilinear-based relative coordinate system to project the actors' motions and trajectories, as can be seen from lines $1-8$ in algorithm \ref{alg:refine_trajectories}.

\textbf{The Refinement Layer} takes in the trajectory set $\Gamma_{set} = \{T_{0}, T_{1}, \ldots, T_{n-1}\}$ and prunes them based on a trajectory's plausibility of existence derived from the HD map. 
Using prior knowledge about the lane widths from the Lane centerlines, we can delineate the boundary regions, allowing us to regress the drivable regions as polygons. This construction results in boundary envelope polygons (blue lines, as can be seen from Fig. \ref{fig:local_buffer_construction}) surrounding the outermost centerline, which can then be used to generate upper and lower sampling bounds for trajectories. A trajectory $T_{i}$ is plausible and considered valid if and only if all points in $T_{i}$ exists inside the boundary polygon. A detailed representation of the refinement layer workings can be seen from the following algorithm \ref{alg:refine_trajectories}. It is essential to note that the refinement process exclusively considers road topology and boundaries, and disregards road occupancy by other actors. The interaction between actors is addressed in the learning phase of the approach.

\begin{algorithm}
\caption{Refinement Layer }
\label{alg:refine_trajectories}
\begin{algorithmic}[1]
    \REQUIRE Trajectory set $\Gamma_{set}$, HD-map $M$
\ENSURE $Feasible\,trajectories \,\Gamma_f \leftarrow$ Local coordinate system

\STATE $origin \leftarrow$ Query origin of the target actor $a_{tar}$
\STATE $tangent \leftarrow$ Query tangent vector of the nearest lane $l$ to $a_{tar}$
\STATE $\theta \leftarrow$ Calculate rotation angle from the tangent vector $tangent$
\STATE $R \leftarrow$ Rotation matrix with rotation angle $\theta$

\FOR{each actor $i$ in scene}
    \STATE Calculate displacement offsets $dx, dy$ for waypoints defining the trajectories
    \STATE $\Tilde{dx},\Tilde{dy} \leftarrow (dx, dy) \cdot R$ \COMMENT{Rotation into curvilinear coordinates}
\ENDFOR

\STATE $Goal_{lanes} \leftarrow$ Query centerline points of nearby lanes to find reachable goal positions.
\STATE $Polygon_{b} \leftarrow$ Construct buffer polygon boundaries from a given lane width from a given lane centerline \COMMENT{In the city coordinate system}
\STATE $\Gamma_{set} \leftarrow \Gamma_{set} \cdot R$ 
\FOR{each trajectory $T_{i} \in \Gamma_{set}$}
    \FOR{each polygon envelope in $Polygon_{b}$}
        \IF{$T_{i}\cap Polygon_{b}\ne \emptyset$}
        \STATE $\Gamma_{set} \setminus T_i$
        \ELSE
        \STATE Continue
        \ENDIF
    \ENDFOR
\ENDFOR

\STATE $\Gamma_f \leftarrow \Gamma_{set} \cdot R$ \COMMENT{Rotation to local coordinate system}
\RETURN $\Gamma_f$, $Goal_{lanes}$
\end{algorithmic}
\end{algorithm}

\textbf{Target Lane Approximations} or goal lanes are pivotal for instructing networks on the “where to go” aspect within a scene. These goal lanes are determined through a straightforward beam search on the lane centerlines, considering a 100-meter radius relative to the heading angle of the target actor. The resulting potential goal lanes are then fed into the Lane Encoder segment of the network to extract pertinent features. This process is essential for the network to learn the “which” information, selecting the most suitable trajectory that aligns with the “where” embeddings.

\begin{figure*}[htb]
    \centering
    \includegraphics[width=0.80\textwidth, keepaspectratio]{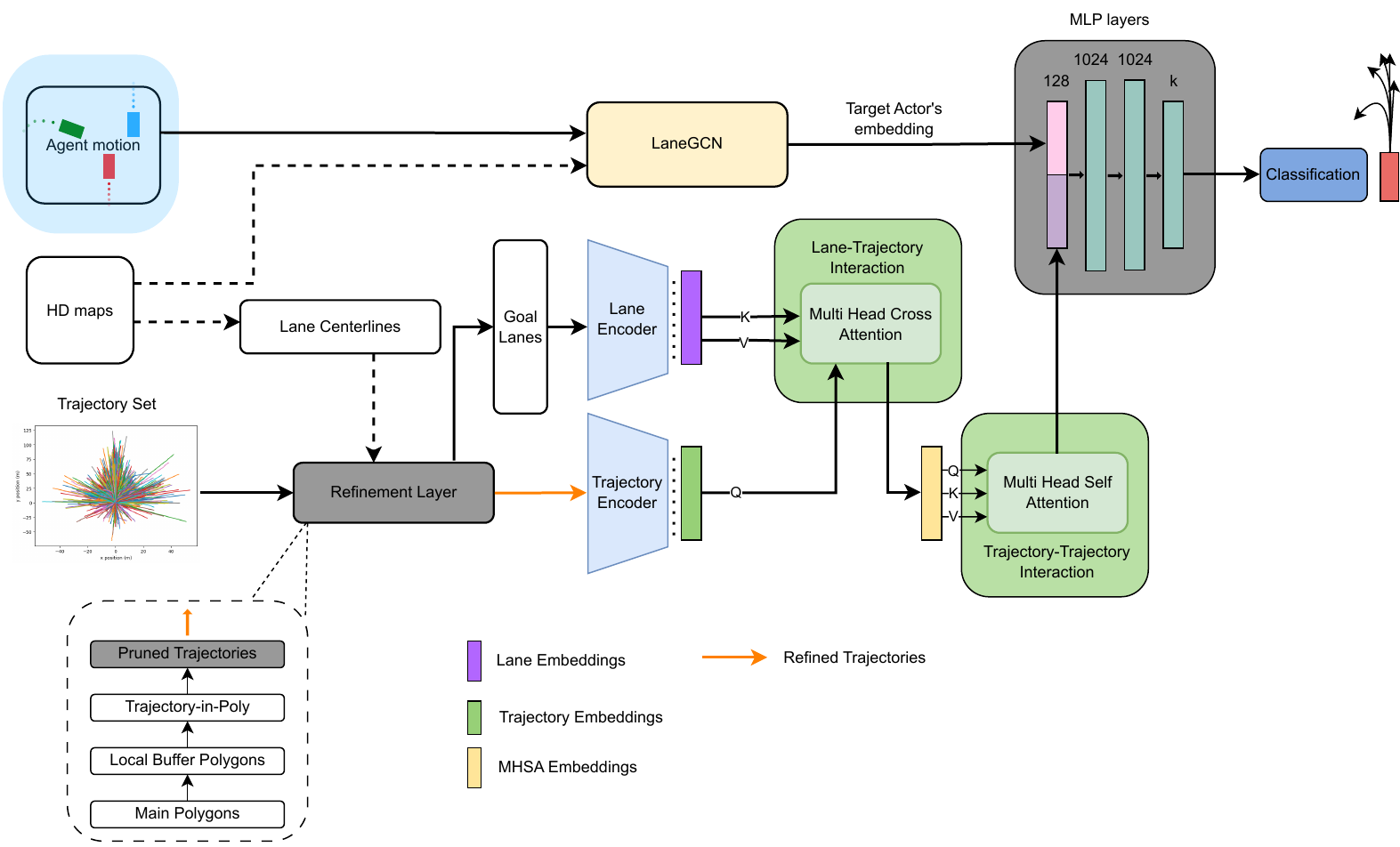}
    \caption{High-level overview of the architecture. The Refinement layer takes two inputs. 1. Lane centerline points in the map coordinate system for a given query window (in meters) and 2. Trajectory sets computed from the dataset with a $\epsilon = 2$ coverage. The refinement layer produces feasible/pruned trajectories by constructing a lane boundary given the lane-centerline points. It also produces a list of possible goal positions or reachable lanes where the target actor could reach within a given prediction horizon. LaneGCN is used as a backbone architecture and outputs an embedding representation for $a_{tar}$ which is then concatenated with the lower parts of the network. }
    \label{fig:kimpf}
\end{figure*}

\subsection{Encoder}
Given $Goal_{lanes}$ and feasible trajectories generated by the refinement layer, we propose learned layers to consume the information and project it into a high-dimensional vector space. Furthermore, we propose the utilization of attention layers to learn interactions between trajectories and lanes. Finally, the encoded feature vector of the trajectories can be extended by the target actor features to enrich its information before passing the final trajectories' representation to a classification head, as shown in Fig.~\ref{fig:kimpf}.

\subsubsection{Learning Goal lanes representation}
$Goal_{lanes}$ is a set of lanes, generated by the refinement layer, that the target actor can traverse given its current heading direction.
A goal lane $L_i \in Goal_{lanes}$ is an ordered set of centerline points on the lane given the direction of the actor's motion, such that $L_i \in \mathbb{R}^{n_i \times 2}$. Each scene can have several goal lanes $R$, and each lane $L_i$ can have a different number of centerline points $n_i$ as a result of differences in road topology and reachabilities. Thus, the goal lanes are dynamically padded during pre-processing such that all lanes in the same scene have the same length $N$. Furthermore, each goal lane is extended with an additional flag set to zero for locations in the lane that are padded. Accordingly, the goal lanes are represented as a set $Goal_{lanes} \in \mathbb{R}^{R \times N \times 3}$. 
The Lane Encoder utilizes temporal convolution and bidirectional LSTM to learn representations of the input $Goal_{lanes}$. 

\subsubsection{Learning feasible trajectories' representation}
A trajectory that is considered feasible is one that can be achieved from the vehicle's initial state, while conforming to the constraints imposed by the road layout and the vehicle's physical capabilities. The Trajectory Encoder is tasked with processing a collection of such feasible trajectories, denoted as, $\Gamma_f \in \mathbb{R}^{D \times t_h \times 2}$ and independently learning a high-dimensional representation for each trajectory. This is accomplished through the application of a temporal convolution followed by an LSTM network.  
\subsection{Attention Layer}
In order to model the interactions between $Goal_{lanes}$ and feasible trajectories $\Gamma_f$ for, $a_{tar}$ we employ the attention mechanism as described in \cite{vaswani2017attention}. 
The multi-head cross-attention layer is particularly adept at integrating two distinct sequences of input, which in this context, facilitates a more profound understanding of how the actor's potential paths (selection of optimal lanes) correlate with the optimal trajectory for execution. The Lane-Trajectory interaction block leverages both trajectory and lane embeddings to execute cross-attention.
In this process, trajectory embeddings are linearly projected to serve as queries (Q), while lane embeddings are used as keys (K) and values (V), following the formulation presented in the attention equation \eqref{eq:attention}. The embeddings that result from this interaction block are then input into the Trajectory-Trajectory interaction block, which is based on self-attention. The multi-head self-attention (MHSA) layer is instrumental in providing the network with the ability to discern global trajectory patterns, thereby enabling trajectories to share information and refine their mutual understanding of the set's inter-trajectory relationships.
For computational efficiency, we use multi-head attention in both these blocks with eight heads. 
\begin{equation}
    Attention(Q, K, V) = softmax(\frac{Q K^T}{\sqrt{d_k}})V,
    \label{eq:attention}
\end{equation}

\subsubsection{Representations fusion and network output}
We employ LaneGCN as the foundational architecture which comprises MapNet, ActorNet, and FusionNet to extract motion representation from actors and map relevant features from the scene. \textit{MapNet}, a graph-based neural network, is tasked with processing the high-definition (HD) map vector representation to derive a high-dimensional embedding of the map. Simultaneously, \textit{ActorNet} encodes the historical trajectories and initial states of the target actor, $a_{tar}$, and other actors, $a_{others}$. ActorNet is designed as a multiscale one-dimensional convolutional neural network (1D CNN) that incorporates a Feature Pyramid Network (FPN) as described by Lin et al. \cite{lin2017feature}. The FPN is instrumental in integrating features extracted at various scales, enhancing the representational capacity of the network.

\textit{FusionNet}, which is an attention-based neural network, leverages the embeddings of the map and actors to capture the interactive dynamics between the map features and the actors within the scene. The output feature representation of the target actor, $a_{tar}$, which is the final product of FusionNet, is then concatenated with the embedding of feasible trajectories obtained from the Trajectory-Trajectory interaction block. This concatenation process is pivotal as it incorporates the interaction information between $a_{tar}$ and $a_{others}$, thereby enriching the trajectory representations with contextual scene interactions.

The combined representation undergoes further fusion through a multilayer perceptron (MLP) layer. This step is crucial for integrating the diverse feature sets into a unified representation suitable for the subsequent classification task. The refined trajectory representation is then input into the classification head, which is responsible for learning the probability distribution over the feasible trajectories.

The classification head is composed of a sequential arrangement of modules, starting with a linear residual block, as proposed by He et al. \cite{he2016deep}, which preserves the dimensionality of the input features. This is followed by a linear layer that adjusts the output dimensionality to match the number of feasible trajectories. Consequently, the network output, denoted as $P$, is a vector of dimensionality $D$, where $P \in \mathbb{R}^D$. Each element within this vector provides a probabilistic estimate of the likelihood of each corresponding trajectory occurring.

\subsection{Loss Learning and Trajectory Scoring}
The classification loss of our network is formulated based on a loss function that has been previously proposed  in the literature \cite{song2022learning, zhao2021tnt}. This loss function employs cross-entropy to quantify the distance between feasible trajectories, denoted as $\Gamma_f$, and the ground truth trajectory, represented as $\Gamma_G \in \mathbb{R}^{1 \times t_h \times 2}$. The essence of this approach is to incentivize the network to prefer feasible trajectories that are proximal to the ground truth trajectory by assigning them higher probabilities, $P$, which are estimated by the network. This mechanism is pivotal for enhancing the accuracy of trajectory prediction.

The assignment of probabilities to feasible trajectories is a critical step in this process. It is achieved by evaluating the Euclidean distance between each feasible trajectory and the ground truth trajectory. The probability associated with the $i^{th}$ feasible trajectory, $T_{f,i}$, is computed using the following formula:
$$
\psi(T_{f,i}, \Gamma_G) = \frac{\exp(-\text{Dist}(T_{f,i}, \Gamma_G) / \tau )}{\sum_{j=1}^{k}\exp(-\text{Dist}(T_{f,j}, \Gamma_G) / \tau)},
$$

where $\tau$ is a hyperparameter known as the temperature factor. This parameter plays a crucial role in controlling the sharpness of the probability distribution over feasible trajectories. The function $\text{Dist}(.)$ calculates the maximum Euclidean distance between two trajectories, considering their respective coordinates at each time step. This distance measure is defined as follows:
$$
\text{Dist}(T_{f,i}, \Gamma_G) = \max_{t} \sqrt{(x_{f,i}^t - x_{G}^t)^2 + (y_{f,i}^t - y_{G}^t)^2},
$$

where $x_{f,i}^t$ and $y_{f,i}^t$ denote the coordinates of the $i^{th}$ feasible trajectory at time $t$, and $x_{G}^t$ and $y_{G}^t$ represent the coordinates of the ground truth trajectory at the same time step.

Subsequently, the network's output, $P$, is processed through a softmax activation function to derive the predicted probabilities, $\gamma(\Gamma_f)$. The classification loss is then determined by computing the cross-entropy between these predicted probabilities and the target probabilities, $\psi(\Gamma_f, \Gamma_G)$. This computation serves as a guiding mechanism for the network, steering it towards generating more accurate trajectory predictions by minimizing the loss.

\textbf{Trajectory selection} Additionally, top-k selection of feasible trajectories is applied in the loss calculation to avoid exhaustive backpropagation of the loss over all the feasible trajectories. In the loss calculation, the top-k trajectories are the $k_{top}$ trajectories with the highest ground truth probabilities, requiring the network to learn a better classification performance over the most similar trajectories to the ground truth. It should be clarified that in the network evaluation phase, the top-k trajectories for computing the evaluation metrics are computed based on the top-k predicted probabilities by the network and have no access to the ground truth information at all. Thus, avoiding injecting any bias or ground truth information into the evaluation process.

\section{Experiments}


\begin{table*}[t]
\centering
\vspace{0.2cm}
\caption{Quantitative comparison of different metrics on Argoverse validation split. Reg; defines the regression-based methods for final trajectory prediction}
\begin{tabularx}{1.80\columnwidth}{lXXXXXXXXXX}
\hline
\centering

Models                                                 & T-set & LowerBound &        &        & K=1    &      &        & K=6    &      & DAC  \\ \hline
                                                       &             & minADE     & minFDE & minADE & minFDE & MR   & minADE & minFDE & MR   &      \\ \hline
Argo-NN+Map (prune) \cite{chang_argoverse_2019}        & —           & Reg        & Reg    & 3.38   & 7.62   & 0.86 & 1.68   & 3.19   & 0.52 & 0.94 \\
Argo-NN \cite{chang_argoverse_2019}                    & —           & Reg        & Reg    & 3.45   & 7.88   & 0.87 & 1.71   & 3.29   & 0.54 & 0.87 \\
Argo-CV \cite{chang_argoverse_2019}                    & —           & Reg        & Reg    & 3.53   & 7.89   & 0.83 & -      & -      & -    & 0.88 \\ \hline
WIMP \cite{khandelwal2020if}                           & —           & Reg        & Reg    & 1.43   & 6.37   & -    & 1.07   & \textbf{1.61}   & 0.23 & -    \\
LaneGCN \cite{lanegcn}                                 & —           & Reg        & Reg    & 1.71   & 3.78   & \textbf{0.67} & \textbf{0.90}   & 1.77   & 0.26 & -    \\
CoverNet \cite{covernet}                               &$\epsilon=2$ & 0.92 & 0.87   & 3.42   & 7.65   & 0.79 & 1.82   & 3.59   & 0.48 & -    \\ \hline
\textbf{Ours(w/ Pre-trained weights)}                  &$\epsilon=2$ & 0.92 & 0.87   & 2.26   & 4.48   & 0.77 & 1.79   & 2.57   & 0.31 & \textbf{0.99} \\
\textbf{Ours(Full)}                                    &$\epsilon=2$ & 0.92 & 0.87   & 2.02   & 4.07   & 0.74 & 1.76   & 2.35   & 0.27 & \textbf{0.99} \\ \hline
\textbf{Ours(w/ Pre-trained weights)}                  &AV1$_{2k}$   & 0.79  & 0.61   & 1.85   & 3.91   & 0.72 & 1.42   & 2.09   & 0.25 & \textbf{0.99} \\
\textbf{Ours(Full)}                                    &AV1$_{2k}$   & 0.79  & 0.61   & \textbf{1.62}   & \textbf{3.41}   & 0.71 & 1.12   & 1.87   & \textbf{0.21} & \textbf{0.99}\\ \hline

\end{tabularx}
\label{tab:table_results}
\end{table*}

\subsection{Dataset}
We utilize the Argoverse-1 motion forecasting dataset, as detailed by Chang et al. \cite{chang_argoverse_2019}. This dataset encompasses 205,942 scenes for training purposes and an additional 39,472 scenes for validation. Each scene is characterized by a target actor, whose past motion is consistently observable throughout the scene. Furthermore, each scene includes additional actors, whose motions may be either partially or fully observable. For other actors denoted as $a_{other}$, where observations are incomplete, we employ zero padding to ensure their shape aligns with that of the target actor. 
The primary objective of our network is to forecast the future trajectory of the target actor, based on past observations of both the target actor and other actors within the scene. The observation horizon is set at $2s$, while the prediction horizon is $3s$, with both horizons sampled at $10Hz$.

\subsection{Trajectory Set Generation}
In this work, a simplified bagging algorithm akin to CoverNet is employed to generate a set of trajectories with a coverage parameter $\epsilon=2$, resulting in a total of 2800 trajectories. While CoverNet utilizes sets with $\epsilon$ values of 3 and 8, our selection of $\epsilon=2$ provides sufficiently dense coverage to validate the efficacy of our proposed method. We also use a metric driven way of set generation proposed in \cite{schmidt2023reset} to validate the experimental results. 

Trajectories are subjected to interpolation and translation to fit to the local coordinate system. Additionally, we compute a theoretical lower bound for the network's performance using the trajectory sets. This lower bound is derived from the minimum Average Displacement Error (minADE) between the ground truth trajectory and the nearest trajectory within the set, as illustrated in Table \ref{tab:table_results}.

\subsection{Metrics}
We evaluate our approach using standard metrics employed in prior studies, namely the minimum Average Displacement Error (minADE), minimum Final Displacement Error (minFDE), and Miss Rate (MR) for both single-modal (k = 1) and multi-modal (k = 6) predictions. Along with these standard metrics, we provide evaluations for drivable area compliance (DAC) \cite{chang_argoverse_2019}. For a model that produces $a$ trajectories out of which $b$ trajectories violate road boundary constraints, the DAC for the model results in $(a-b)/a$. A higher DAC indicates higher compliance with the drivable area. 

\subsection{Training details}
The network is trained for $38$ epochs with Adam optimizer \cite{kingma2014adam} on an NVIDIA A100 GPU. The optimizer has a learning rate of $10^{-3}$ for the first 30 epochs and changes to  $10^{-4}$ for the remaining 8 epochs; with a batch size of $32$. All layers in the architecture have a dimension of $128$. The number of heads in the attention layer is $8$. The temperature factor $\tau$ and $k_{top}$ are set to $1$ and $6$ in the loss calculation.

\subsection{Computational performance} 
\begin{figure}[h]
    \centering
    \includegraphics[width=0.95\columnwidth]{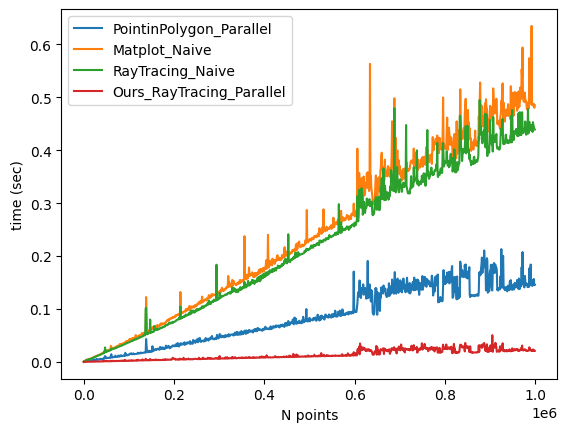}
    \caption{Performance of various methods to employ Point-in-Polygon operation. We use $Ours\_RayTracing\_Parallel$ algorithm in our refinement layer for pruning non-compliant trajectories.}
    \label{fig:computation_graph}
\end{figure}

Although the refinement layer does not have any learnable parameters, it requires computation time to perform the refinement, which is crucial for the real-time deployment of the network. In our refinement layer, we use the point-in-polygon operation to perform the boundary conditions check. On average, per scene, we were able to perform a check under $0.3-0.5 ms$ for $4$ polygon envelops on 12 core CPU. The check could be parallelized easily, allowing for a significantly faster computation on a GPU. Fig.~\ref{fig:computation_graph} showcases the theoretical limits for a point-in-polygon operation, which we performed using different algorithms on a CPU for a total of 1 million randomly sampled points, allowing us to justify that the refinement layer could meet real-time requirements. 

\section{Results and Discussion}

The comparative analysis of our research is conducted against similar methodologies using the Argoverse validation split dataset, as shown in Table \ref{tab:table_results}. To maintain a fair and consistent comparison, we only include methods that have reported results for all $k=1$, $k=6$, and MR. We trained our models with 2 variants of trajectory sets; one using the CoverNet like bagging algorithm ($\epsilon=2$) and the other through Metric-Driven way (AV1$_{2k}$), following the implementations from \cite{schmidt2023reset}. Our primary objective is to prevent off-road predictions, which is reflected in our high Driving Area Compliance (DAC) score of 0.99. This score suggests that our method effectively excludes nearly all unfeasible states from the prediction space utilized by the network. 

By comparing our results with the Covernet and LaneGCN, our model outperforms both the minADE and minFDE aspects for $k=1$, with the AV1$_{2k}$set, whilst performing well against CoverNet on all metrics, given the  lower bounds for our model. Our model performed worse on the Euclidean metrics for $k=6$, when compared with the regression methods (Reg) and this decrease can be attributed to the sparse trajectory set over which our model has to make a classification on.

The MR of 0.21 for our \textit{Ours (Full)} achieved the best result when compared with the State-of-the-art, as this variant was trained from scratch without using pre-trained weights for the backbone, unlike \textit{Ours (w/ Pre-trained weights)}. This suggests that our model trained from scratch focuses on learning more concrete representations of plausible states when presented with future reachable states and goal-oriented lanes, without compromising on the quality of the predictions. 

\subsection{Limitations}
The observed degradation in predictive performance relative to the baseline LaneGCN can be ascribed to the intrinsic properties of the trajectory sets. An empirical examination indicates that nearly $48\%$ of the trajectories are discarded due to the static nature inherent in precomputing such a set. To mitigate this, we are currently devising multiple trajectory sets as an alternative to a single set. These sets can be swapped during runtime, thereby enabling a denser representation that more faithfully represents the true underlying distribution. This strategy is anticipated to augment the representation density and more effectively encapsulate the true underlying distribution, thereby potentially improving the performance on other metrics.

\section{Conclusion}
In this paper, we introduce a novel framework for multi-modal motion forecasting that ensures the physical feasibility of trajectories by incorporating road topology preventing off-road predictions. Our approach follows a two-step modular process. Firstly, a deterministic refinement layer is employed to generate physically reachable trajectories and goal positions. Secondly, a learnable layer captures global interactions among maps, reachable states, and feasible trajectories, providing comparative results to regression-based approaches.

\addtolength{\textheight}{-0cm}   


\section*{ACKNOWLEDGMENT}
The research leading to these results was funded by the German Federal Ministry for Economic Affairs and Climate Action and was partially conducted in the projects “KI Wissen” and “SafeADArchitect”. Responsibility for the information and views set out in this publication lies entirely with the authors.


\printbibliography
 
\end{document}